\documentclass[11pt]{article}
\usepackage{nodalida2025}
\usepackage{times}
\PassOptionsToPackage{hyphens}{url}
\usepackage{array}
\usepackage{amsmath}

\usepackage[table,xcdraw]{xcolor}
\definecolor{darkblue}{rgb}{0, 0, 0.5}
\usepackage[colorlinks=true, citecolor=darkblue, linkcolor=darkblue, urlcolor=darkblue]{hyperref}
\usepackage{url}            
\usepackage{booktabs}       
\usepackage{nicefrac}       
\usepackage{appendix}
\usepackage{multirow}
\usepackage{boldline}
\usepackage{algorithm}  
\usepackage{algpseudocode}  

\usepackage{pgfplots}
\usepackage{pgfplotstable}
\pgfplotsset{compat=1.18}
\usepgfplotslibrary{groupplots}

\usepackage{tikz}

\aclfinalcopy 

\newcommand{\mimir}{Mímir}

\title{The Impact of Copyrighted Material on  Large Language Models:\newline A Norwegian Perspective}

\author{
Javier de la Rosa$^1$
Vladislav Mikhailov$^2$
Lemei Zhang$^3$
Freddy Wetjen$^1$
David Samuel$^2$
\AND
Peng Liu$^3$
Rolv-Arild Braaten$^1$
Petter Mæhlum$^2$
Magnus Breder Birkenes$^1$
\AND
Andrey Kutuzov$^2$
Tita Enstad$^1$
Hans Christian Farsethås$^2$
Svein Arne Brygfjeld$^1$
\AND
Jon Atle Gulla$^3$
Stephan Oepen$^2$
Erik Velldal$^2$
Wilfred Østgulen$^1$
Lilja Øvrelid$^2$
\AND
Aslak Sira Myhre$^1$
\\ \\
$^1$National Library of Norway \\
$^2$University of Oslo \\
$^3$Norwegian University of Science and Technology
\\ \\ \small{
    \textbf{Correspondence:} {\texttt{versae@nb.no}}
}}

\date{}

\begin{document}
\maketitle
\begin{abstract}
The use of copyrighted materials in training language models raises critical legal and ethical questions. This paper presents a framework for and the results of empirically assessing the impact of publisher-controlled copyrighted corpora on the performance of generative large language models (LLMs) for Norwegian. When evaluated on a diverse set of tasks, we found that adding both books and newspapers to the data mixture of LLMs tend to improve their performance, while the addition of fiction works seems to be detrimental. Our experiments could inform the creation of a compensation scheme for authors whose works contribute to AI development.

\end{abstract}

\section{Introduction}
Generative language models have radically reshaped the landscape of natural language processing (NLP), enabling the development of systems that can generate and interact with human language at an unprecedented level. This includes Norwegian, for which several large language models (LLMs) have been trained and published in the recent years using different architectures and licensing choices \cite{kummervold-2021-nb-bert,kutuzov-etal-2021-large,samuel-etal-2023-norbench,samuel2025small,norglm}.

However, the vast quantities of data required for training these models often include copyrighted materials, presenting novel challenges related to intellectual property rights and compensation. Additionally, prior research has highlighted significant concerns about dataset composition and quality in large-scale web-crawled datasets, emphasizing the need for more responsible data curation practices \citep{kreutzer-etal-2022-quality, artetxe-etal-2022-corpus, penedo2024fineweb}. Together, these challenges have led to numerous lawsuits across jurisdictions, fundamentally questioning the legitimacy of training models on copyrighted data without explicit permissions from content creators \citep{Panettieri2024, Madigan2024, bakerlaw2024case}.\footnote{See \citet{gervais2024copyright} for an in-depth introduction on how LLMs are being interpreted in the legal domain.}
 
The first wave of lawsuits emerged shortly after the public release of advanced generative AI models (see Appendix \ref{app:legal_cases}). Content creators, including authors, visual artists, and musicians, began to express concerns about the unauthorized use of their work in training datasets. Multiple class-action lawsuits were filed in the United States, accusing prominent AI companies such as OpenAI and Meta Platforms of infringing on copyright laws by using copyrighted materials without obtaining explicit permissions. The authors argued that the unauthorized use of their works without any form of compensation or recognition undermines their intellectual property rights and jeopardizes their ability to earn a living from their creative endeavors. In Europe, a coalition of news publishers has taken legal action against Google and Meta Platforms, arguing that the use of journalistic content in training models without fair remuneration constitutes a breach of copyright and undermines the sustainability of high-quality journalism. Likewise, Norwegian rights-holder organizations representing publishing houses across the country, contacted the government in late 2023 expressing their concerns over the use of their material in training generative language models and demanding some sort of compensation were their contents to be used in the training of generative language models. As a result, the Ministry of Culture and Equality (\textit{Kultur- og likestillingsdepartemente}) instructed the National Library to create a data-driven report they could use in order to make informed decisions in the elaboration of a compensation scheme for the authors. Led by the National Library of Norway, a consortium was formed together with the University of Oslo and the Norwegian University of Science and Technology under the umbrella of the so-called \mimir{} Project.\footnote{A name chosen after a figure in Norse mythology renowned for his knowledge and wisdom.}




In this context, and under the umbrella of \mimir, this paper describes a first attempt at empirically evaluating the impact of copyrighted content in the training of LLMs for Norwegian. We introduce a set of carefully curated datasets that are later used in the training of foundational, domain-tuned, and instruction-tuned models. We establish the proper training conditions to be able to compare models trained on the different datasets. A newly created benchmarking suite is used to evaluate the performance of each individual model and make the comparison meaningful. As a collaborative effort among several institutions, the results of our investigations set the basis to guide policymaking and proper compensation schemes for authors and right-holders in Norway \citep{mimir2024project}.




\section{Methodology}

The methodology involves a comprehensive analysis that spans several stages. Initially, a diverse corpus of primarily Norwegian language data is assembled, incorporating both copyrighted and non-copyrighted materials, plus materials commonly found on the Internet. This corpus serves as the foundation for training various LLMs, each with different configurations and access levels to copyrighted content. By comparing the performance of these models across a range of linguistic and natural language processing tasks, such as text generation, translation, summarization, question-answering, sentiment analysis and more, we seek to quantify the specific contributions of copyrighted materials to the overall model quality.

To ensure robustness and reliability, the evaluation framework focuses on generation capabilities, natural language understanding, and linguistically-inspired metrics. Quantitative measures include traditional NLP metrics like accuracy, F1, BLEU, and ROUGE, which provide assessments of model accuracy and fluency. Linguistic analysis, on the other hand, involves assessing the coherence, language variability, and contextual relevance of the generated outputs. This dual approach allows for a nuanced understanding of how copyrighted materials impact the performance and utility of LLMs.

\section{Data Collection}

With the objective of setting up a realistic training scenario where using Internet crawled sources is commonplace, we gathered publicly available text collections like Wikipedia, datasets from the HPLT \cite{de-gibert-etal-2024-new} and CulturaX \cite{nguyen-2024-culturax} projects, code in different programming languages from \citet{lozhkov2024starcoder}, governmental reports and publications published under open licenses, and books and newspapers articles in the public domain.

We then collaborated with the National Library of Norway and the rights-holder organizations to gain access to protected materials. Through the legal deposit act, the National Library of Norway has digitized almost all books in Norwegian and around 85\% of the newspapers ever published in the country \cite{nasjonalbiblioteket_arsrapportar}. Where the quality of the digitized material was not enough (e.g., due to OCR processing), or was not been legally deposited (e.g., paywalled content), specific agreements were put in place to obtain the material from third party organizations such as the Norwegian Broadcasting Corporation (NRK), the TV channel TV2, and the newspaper conglomerates Amedia and Schibsted. In line with provisions that allow research on language technology and data mining (\textit{Åndsverkloven}), and with the consent of the Norwegian right-holders, this study primarily relied on material legally deposited at, or under agreement with, the National Library of Norway. Specifically, we focus our study on the collection of publisher-controlled books and newspapers articles.

\subsection{Core Datasets}
\begin{figure*}
    \centering
    \includegraphics[width=0.98\linewidth]{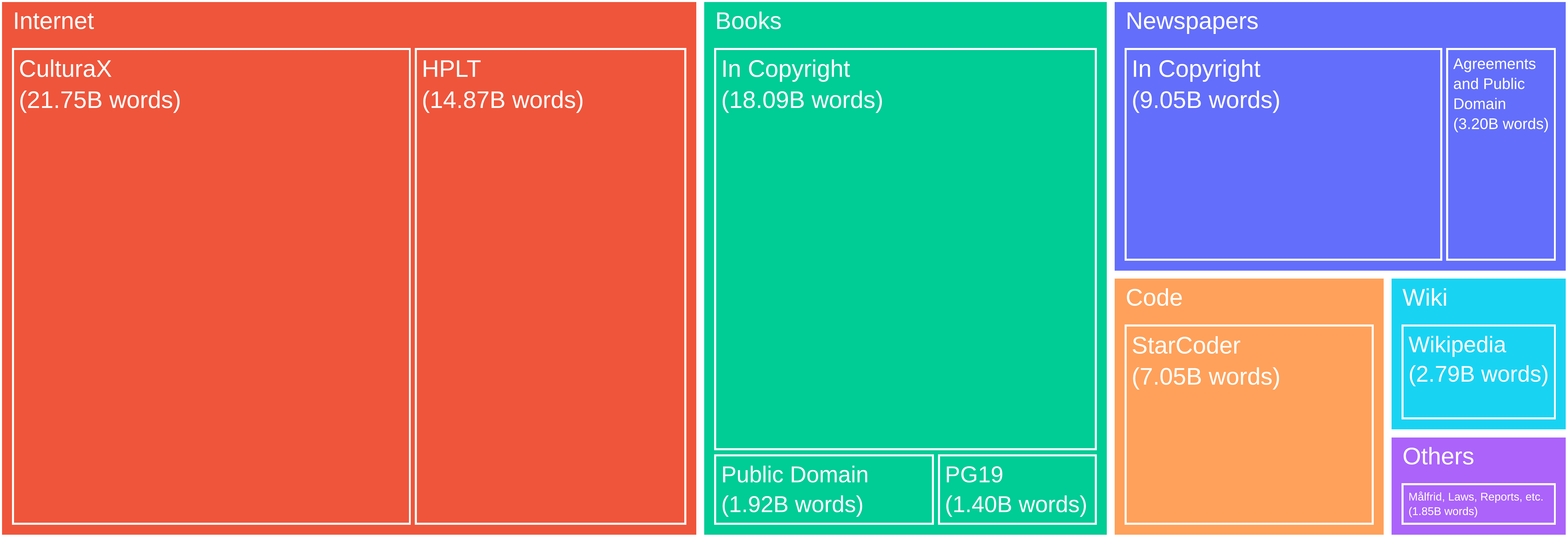}
    \caption{Treemap with the final number of words (comma separated) contributed by each source after cleaning and deduplication.}
    \label{fig:treemap}
\end{figure*}

This mixture of data (see Figure \ref{fig:treemap} and Appendix \ref{app:sources}) allowed us to evaluate the impact of high-quality publisher-controlled copyright-protected corpora versus other sources commonly available on the Internet. The models trained on the copyrighted materials will not be made publicly available for further use and only serve the purpose of this study.


We followed the recipe of the Norwegian Colossal Corpus (NCC) by \citet{kummervold-2022-ncc}, adapting and updating it with new up-to-date contents, re-OCRing some materials, enriching their metadata, and ensuring uniform format and functionality across datasets. The preparation involved cleaning, deduplication, metadata tagging, and language balancing to maintain consistent representation of Norwegian, preventing other languages from overshadowing it. The corpus was divided into two main datasets: a \textbf{base} dataset excluding publisher-controlled copyright-protected books and newspapers,\footnote{Except for newspapers that fall under the Language Technology Use (\textit{Språkteknologiformål}), as they were already included in other datasets such as NCC.} and an \textbf{extended} dataset that included all collected texts, thus including all of \textbf{base} (see Table \ref{tab:sources}).

\begin{table}
    \centering
    \begin{tabular}{lrr}
    \textbf{Dataset} & \textbf{Documents} & \textbf{Words} \\
    \midrule
    base & 60,182,586 & 40,125,975,241\\
    extended & 125,285,547 & 82,149,281,266  \\
    \end{tabular}
    \caption{Number of documents and words in each of the core datasets. Words refer to whitespace-separated sub-strings.}
    \label{tab:sources}
\end{table}

We decided to include texts from other Scandinavian (Swedish, Danish, and Icelandic) and English sources to boost the performance of the resulting language models via cross-lingual transfer \citep{conneau-etal-2020-emerging, xue-etal-2021-mt5}. To ensure that languages other than Norwegian, and primarily coming via Internet crawling, were balanced, we adapted the perplexity-based sampling strategy from \citet{bertin-delarosa-2022} to maintain a high quality in the selected data. Instead of sampling a fixed number of documents, parameters for a Gaussian curve were calculated from 500,000-1M random documents per source, utilizing Wikipedia-based Kneser-Ney language models from \citet{wenzek2019ccnet} and \citet{conneau2019unsupervised}.\footnote{Built with KenLM \cite{heafield2011kenlm}.} We also modified the perplexity calculation to account for normalized text. These parameters then guided dataset sub-sampling to target ratios per language, reducing foreign language content while maintaining quality (Appendix \ref{app:sampling}).

It is also important to notice than in order to maintain the language distributions for foreign languages with respect to the amount of Norwegian texts, the total number of documents in foreign languages in the \textbf{extended} dataset is consequently higher and slightly different (due to the sampling strategies) than that of \textbf{base}; we keep the same ratios (see Appendix  \ref{app:sources}).

\subsection{Domain Specific Subsets}
\begin{table}
    \centering
\resizebox{8cm}{!}{
    \begin{tabular}{lrr}
        \textbf{Subset} & \textbf{Documents}  & \textbf{Words} \\
        \midrule
        books & 492,281 & 18,122,699,498 \\
        newspapers & 46,764,024  & 9,001,803,515 \\
        books + newspapers & 47,256,305  & 26,078,915,554 \\
       \midrule
       fiction books  & 117,319  & 5,287,109,366 \\
       nonfiction books  & 359,979  & 12,384,323,012 \\
       nonfiction books + newspapers  & 42,083,532  & 20,340,539,068 \\
       \midrule
       original books  & 392,887  & 13,352,261,605 \\
       original books + newspapers  & 47,156,911  & 22,354,065,120 \\
       translated books  & 96,258  & 4,695,814,506 \\
    \end{tabular}
}
    \caption{Number of documents and words (comma separated) in each subset of the publisher-controlled corpora.}
    \label{tab:subsets}
\end{table}

The publisher-controlled copyright-protected materials present in the \textbf{extended} dataset were further divided into groups attending to different criteria. These subsets were carefully designed to test the effect of adding them to the training sets for LLMs. We split the books into fiction vs nonfiction, and original works in Norwegian vs translations. While most books in the collection had metadata information regarding the original language in which a work was written in, genre labels were more scarce. To overcome this limitation, we built a Doc2Vec model \cite{le-mikolov2014doc2vec} that classified fiction vs nonfiction with 98\% accuracy and used it to annotate books for which this information was missing.\footnote{\url{https://huggingface.co/mimir-project/literary-form-classifier}} As shown in Table \ref{tab:subsets}, we then built domain specific subsets to investigate 1) the effect of books vs newspapers vs books + newspapers, 2) the effect of factuality by adding only fiction words, only nonfiction works, and nonfiction works + newspapers, and 3) the effect of adding content written originally in Norwegian, such as original books or original books + newspapers, vs translated books.

\subsection{Instruction-tuning Datasets}
To align the models more closely with human objectives and assess whether instruction tuning with limited high-quality data can enhance the performance of our pre-trained models across various tasks, we built upon prior work and collected nearly 5,000 instructions annotated by research assistants.\footnote{\url{https://huggingface.co/datasets/mimir-project/mimir-instruction}} The instructions were formatted as \textit{(instruction, input, output)} triplets, where \textit{instruction} refers to the directive provided by humans for the model, \textit{input} is an optional field containing task-related information, and \textit{output} denotes the desired response that follows the given instruction.

The instruction tuning dataset combines three key categories --Reading Comprehension, Norwegian Culture, and Words and Expressions-- with diverse domains to enhance model performance. The domains include Literature, Commonsense, Geography, Language, History, Sports, Entertainment, Food, Politics, Science, Art, Music, and Culture. The variety of the instructions seeks to improve the model’s ability to understand complex texts, provide culturally relevant responses, and handle language nuances, resulting in more versatile, knowledgeable, and context-aware LLMs.

\begin{table*}
  \footnotesize
  \centering
  \begin{tabular}{llrr}
    \textbf{Model} & \textbf{Initialization} &  \textbf{GPU/hours} & \textbf{Accelerator} \\
    \hline
    \rowcolor{lightgray!50} \multicolumn{4}{c}{Core Models}\\ \hline
    base & From scratch &  50K & AMD MI250X \\
    extended & From scratch &  50K & AMD MI250X \\
    base (warm) & Mistral 7B v0.1 &  13.8K & NVIDIA H100 \\
    extended (warm) & Mistral 7B v0.1 &  55.6K & AMD MI250X  \\
    \hline
    \rowcolor{lightgray!50} \multicolumn{4}{c}{Domain Tuned Models}\\ \hline
    base + fiction books & base &  7.5K  & AMD MI250X \\
    base + nonfiction books & base &  7.5K & AMD MI250X \\
    base + nonfiction books + newspapers & base &  7.5K & AMD MI250X \\
    base + newspapers & base &  4.8K & Google TPUv4 \\
    base + books & base &  4.8K & Google TPUv4 \\
    base + books + newspapers & base &  4.8K & Google TPUv4 \\
    base + original books + newspapers & base &  9.1K & AMD MI250X \\
    base + original books & base &  9.1K & AMD MI250X \\
    base + translated books & base &  9.1K & AMD MI250X \\
    \hline
    \rowcolor{lightgray!50} \multicolumn{4}{c}{Instruction Fine Tuned Models}\\ \hline
    base \textit{instruct} & base &  14.2 & NVIDIA H100 \\
    extended \textit{instruct} & extended &  14.2 & NVIDIA H100 \\
    base (warm) \textit{instruct} & base (warm) &  14.2 & NVIDIA H100 \\
    extended (warm) \textit{instruct} & extended (warm) & 14.2 & NVIDIA H100 \\
  \end{tabular}
  \caption{Model training specifications, where \textit{Model} represents the model identifier and the data used for training, \textit{Initialization} represents the base model used for training, \textit{GPU/hours} indicates the total GPU hours required for model training, and \textit{Accelerator} represents the type of accelerator used.}
  \label{tab:model_list}
\end{table*}

\section{Model Training}
The training phase involved multiple models, each based on the Mistral architecture \cite{jiang2023mistral7b}. The training was conducted in the following stages.

\begin{enumerate}\itemsep0em
    \item To measure the overall impact of publisher-controlled copyrighted corpora and its effect in realistic scenarios, we conducted pre-training on the \textbf{base} and \textbf{extended} datasets, both from scratch and using the existing weights (warm) of the pre-trained model Mistral 7B v0.1.\footnote{\url{https://huggingface.co/mistralai/Mistral-7B-v0.1}} These four \textit{core models} were trained on the same amount of data, 64,000 steps of 4 million sub-word tokens each. using identical sets of hyperparameters (see Table \ref{tab:hyperparams} in Appendix \ref{app:hyperparams}). This roughly translates to 3 epochs for the \textbf{base} dataset and 2 for the \textbf{extended} dataset, which according to \citet{muennighoff2023scaling} is still far from saturating the available data.
    \item To further isolate the effect of different ablations of the publisher-controlled copyright-protected corpora, we continuously fine tuned the model trained on \textbf{base} from scratch for an extra 10,000 steps on each of the 9 domain specific subsets. 
    \item The core models were also fine tuned on the instruction data for 4 iterations to evaluate their performance on downstream tasks.
\end{enumerate}

Overall, we trained 17 models (7 billion parameters each) using a total of 270,000 GPU-hours. Model training specifications are shown in Table \ref{tab:model_list}. The infrastructure for training included the LUMI supercomputer, Idunn cluster, and Google
TPUs through the Tensor Research Cloud program\footnote{To assess the deviation introduced by differences in training infrastructures and platforms across the participating institutions, each team trained a control model with 1.5B parameters based on the Llama 2 architecture. The training setups were identical, utilizing the \textbf{base} dataset. After comparing the validation loss curves from each team, we found that the curves were almost the same, with a deviation of less than $0.05$ in terms of the final convergence validation loss.}. Besides, we trained two tokenizers with the \textbf{base} and \textbf{extended} datasets separately, both with the same vocabulary size of $32,768$. After an initial test of the fertility of the tokenizers,\footnote{Fertility expresses the fragmentation rate of a tokenizer and is $\#tokens / \#words$ in one corpus.} we found the difference between them was only $0.0013$. Therefore, we decided to use the same tokenizer trained with the \textbf{base} dataset for all the models.

\section{Evaluation Framework} In our empirical evaluation experiments, we utilize \texttt{NorEval},\footnote{\url{https://github.com/ltgoslo/noreval}} a publicly available framework for evaluating Norwegian generative LLMs built on \texttt{lm-evaluation-harness} \cite{eval-harness}. We consider 28 tasks, which test model's various Norwegian language understanding and generation abilities. \texttt{NorEval} covers both Norwegian language varieties (Bokmål and Nynorsk) and provides a set of 4--6 prompts for each downstream task. The tasks can be grouped into nine higher level \textbf{skills}:

\begin{enumerate}\itemsep0em 
    \item \textbf{Sentiment Analysis}, here defined as binary polarity classification on both the sentence- and document-level based on the existing \texttt{NoReC}
datasets of professional reviews \cite{velldal-etal-2018-norec,ovrelid-etal-2020-fine}.
    \item \textbf{Fairness \& Truthfulness}. Fairness refers to the absence of bias in the predictions and outputs of a model. Evaluating fairness ensures that the model does not favor or discriminate against particular groups based on attributes like race, gender, or ethnicity. This skill was evaluated using a newly-created dataset,\footnote{\url{https://huggingface.co/datasets/mimir-project/mimir-bias}} 
    which covers a wide range of bias types, including race, religion, gender, geography, occupation, age etc. Truthfulness involves the accuracy and reliability of the information produced by the model, ensuring it generates factual and verifiable content. This skill was evaluated using \texttt{NorTruthfulQA} \cite{mikhailov2025collection}, which  assesses whether a model is truthful in selecting and generating answers to questions that involve common human misconceptions.\footnote{\url{https://huggingface.co/datasets/ltg/nortruthfulqa_mc} and \url{https://huggingface.co/datasets/ltg/nortruthfulqa_gen}} 
    \item \textbf{Reading Comprehension}, which measures the ability of a model to understand and interpret text. It involves answering questions about a given passage, summarizing content, or explaining the meaning of specific phrases or sentences. This skill estimates how well the model grasps the context and details in the text. It was evaluated using the existing extractive question-answering \texttt{NorQUAD} dataset \cite{ivanova-etal-2023-norquad} and multiple-choice question-answering Belebele dataset \cite{bandarkar-etal-2024-belebele}.
     \item \textbf{World Knowledge}, which assesses the extent of factual information a language model has about the world. This includes historical events, geographical data, scientific facts, cultural knowledge, and more. The model should correctly answer questions or provide information based on real-world knowledge. This skill was evaluated using the \texttt{NorOpenBookQA} 
       and \texttt{NRK-Quiz-QA} by \citet{mikhailov2025collection}.\footnote{\url{https://huggingface.co/datasets/ltg/noropenbookqa} and \url{https://huggingface.co/datasets/ltg/nrk_quiz_qa}}
   \item \textbf{Commonsense Reasoning}, which involves the ability of a model to make logical inferences based on everyday knowledge and understanding of the world. The model should reason about situations that require practical, everyday knowledge that people take for granted. This skill was evaluated using \texttt{NorCommonsenseQA} \cite{mikhailov2025collection},\footnote{\url{https://huggingface.co/datasets/ltg/norcommonsenseqa}} which consists
       of multiple-choice commonsense question answer-pairs which adapts the corresponding English CommonsenseQA dataset \cite{talmor-etal-2019-commonsenseqa} to Norwegian.
 \item \textbf{Norwegian Language} evaluation focuses on the ability of a model to understand and generate text in Norwegian, specifically its grammar, structure, and sentence construction. This skill is important for assessing how well the model handles Norwegian  and their specific syntactic rules. It was evaluated using the existing \texttt{NCB} \cite{FarsethasTjostheim2024} and \texttt{ASK-GEC} \cite{jentoft2023grammatical} datasets, and the newly-created \texttt{NorIdiom} dataset.\footnote{\url{https://huggingface.co/datasets/mimir-project/noridiom}} 
  \item \textbf{Summarization}, which measures the ability of a model to condense longer pieces of text into shorter, coherent summaries that capture the main points. This skill is crucial for applications where users need a quick understanding of large volumes of information, such as news articles or research papers. It was evaluated using the \texttt{NorSumm} dataset \cite{touilebetal2025}.\footnote{\url{https://huggingface.co/datasets/SamiaT/NorSumm}}
   \item \textbf{Translation}, which assesses how accurately a language model can convert a text from one language to another while preserving the meaning, tone, and context. 
   It was evaluated using the existing \texttt{Tatoeba} dataset \cite{tiedemann-2020-tatoeba}. The following six language pairs are considered: Bokmål $\leftrightarrow$ Nynorsk, Bokmål $\leftrightarrow$ English, and English $\leftrightarrow$ Nynorsk. 
    \item \textbf{Variation and Readability}, which consists of measuring the lexical diversity of a model by looking at the amount of redundancy in the text it produces and at the readability of these texts measured by average sentence length and the proportion of long words. As such, this skill evaluation did not require any specific benchmarking datasets.
\end{enumerate}

We follow the standard in-context learning evaluation design for pretrained decoder-only language models \citep[e.g.,][]{brown2020language,touvron2023llama}, which includes zero-shot and few-shot evaluation.  In this paper, for the sake of simplicity, we selected the most common metrics per task and aggregated scores using a simple cumulative sum per higher-level skill. In order to aggregate results into an overall score, with the caveats of aggregating metrics of different nature, scores were extracted for the best available \{0, 1, 4, 16\}-shot configuration for each task and the best score for each of the prompts. Metrics were normalized to exhibit the same higher-is-better behavior in a range of 0 to 100.

\begin{figure*}
\begin{tikzpicture}
\begin{axis}[
    width=12.5cm,
    height=10cm,
    xbar stacked,
    bar width=12pt,
    xlabel={Scores},
    symbolic y coords={
        Mistral 7B v0.1,
        base (warm),
        extended (warm),
        base + translated books,
        base + original books + newspapers,
        base + original books,
        base + nonfiction books + newspapers,
        base + nonfiction books,
        base + fiction books,
        base + books + newspapers,
        base + newspapers,
        base + books,
        base,
        extended
    },
    ytick=data,
    yticklabel style={font=\footnotesize},
    legend style={
        at={(0.3,0.0)},
        anchor=north,
        legend columns=3,
        cells={anchor=west},
        font=\footnotesize,
        draw=none, 
    },
    xmin=0,
    xmax=550,
    legend image code/.code={
        \draw[#1] (0cm,-0.1cm) rectangle (0.3cm,0.1cm);
    },
    reverse legend=false,
    nodes near coords align={left},
    axis line style={draw=none},
    tick style={draw=none},
    xtick=\empty,
    xlabel={}
]

\draw[black] (rel axis cs:0,0.8205) -- (rel axis cs:1,0.8205);

\draw[black] (rel axis cs:0,0.2435) -- (rel axis cs:1,0.2435);

\draw[black] (rel axis cs:0,0.1155) -- (rel axis cs:1,0.1155);

\addplot[fill=blue!60] coordinates {
    (88.41,{Mistral 7B v0.1})
    (88.17,{base (warm)})
    (86.57,{extended (warm)})
    (72.22,{base + translated books})
    (77.72,{base + original books + newspapers})
    (75.46,{base + original books})
    (78.99,{base + nonfiction books + newspapers})
    (76.30,{base + nonfiction books})
    (74.16,{base + fiction books})
    (78.94,{base + books + newspapers})
    (76.57,{base + newspapers})
    (76.20,{base + books})
    (69.54,base)
    (77.09,extended)
};

\addplot[fill=red!60] coordinates {
    (64.93,{Mistral 7B v0.1})
    (51.28,{base (warm)})
    (54.64,{extended (warm)})
    (48.21,{base + translated books})
    (54.43,{base + original books + newspapers})
    (55.43,{base + original books})
    (53.91,{base + nonfiction books + newspapers})
    (56.51,{base + nonfiction books})
    (47.99,{base + fiction books})
    (53.06,{base + books + newspapers})
    (48.04,{base + newspapers})
    (55.15,{base + books})
    (53.51,base)
    (51.01,extended)
};

\addplot[fill=yellow!60] coordinates {
    (56.68,{Mistral 7B v0.1})
    (52.48,{base (warm)})
    (52.79,{extended (warm)})
    (34.54,{base + translated books})
    (35.42,{base + original books + newspapers})
    (32.87,{base + original books})
    (36.66,{base + nonfiction books + newspapers})
    (31.84,{base + nonfiction books})
    (33.71,{base + fiction books})
    (35.09,{base + books + newspapers})
    (35.29,{base + newspapers})
    (33.38,{base + books})
    (38.04,base)
    (39.38,extended)
};

\addplot[fill=green!60] coordinates {
    (58.86,{Mistral 7B v0.1})
    (53.27,{base (warm)})
    (51.51,{extended (warm)})
    (40.97,{base + translated books})
    (41.71,{base + original books + newspapers})
    (41.56,{base + original books})
    (41.85,{base + nonfiction books + newspapers})
    (41.06,{base + nonfiction books})
    (39.74,{base + fiction books})
    (41.53,{base + books + newspapers})
    (41.55,{base + newspapers})
    (41.54,{base + books})
    (41.10,base)
    (43.40,extended)
};

\addplot[fill=orange!60] coordinates {
    (36.01,{Mistral 7B v0.1})
    (48.02,{base (warm)})
    (49.25,{extended (warm)})
    (43.33,{base + translated books})
    (41.46,{base + original books + newspapers})
    (41.08,{base + original books})
    (42.68,{base + nonfiction books + newspapers})
    (40.66,{base + nonfiction books})
    (41.39,{base + fiction books})
    (41.77,{base + books + newspapers})
    (37.79,{base + newspapers})
    (39.59,{base + books})
    (39.85,base)
    (41.26,extended)
};

\addplot[fill=cyan!60] coordinates {
    (31.49,{Mistral 7B v0.1})
    (49.51,{base (warm)})
    (48.48,{extended (warm)})
    (43.63,{base + translated books})
    (53.81,{base + original books + newspapers})
    (53.04,{base + original books})
    (54.50,{base + nonfiction books + newspapers})
    (52.48,{base + nonfiction books})
    (41.18,{base + fiction books})
    (52.11,{base + books + newspapers})
    (58.16,{base + newspapers})
    (51.69,{base + books})
    (38.28,base)
    (47.64,extended)
};

\addplot[fill=blue!30] coordinates {
    (10.09,{Mistral 7B v0.1})
    (35.88,{base (warm)})
    (36.14,{extended (warm)})
    (27.51,{base + translated books})
    (30.67,{base + original books + newspapers})
    (28.74,{base + original books})
    (30.78,{base + nonfiction books + newspapers})
    (29.52,{base + nonfiction books})
    (28.41,{base + fiction books})
    (31.86,{base + books + newspapers})
    (29.26,{base + newspapers})
    (29.81,{base + books})
    (32.45,base)
    (41.00,extended)
};

\addplot[fill=red!30, nearly opaque] coordinates {
    (41.55,{Mistral 7B v0.1})
    (41.14,{base (warm)})
    (42.48,{extended (warm)})
    (36.97,{base + translated books})
    (40.64,{base + original books + newspapers})
    (38.86,{base + original books})
    (40.50,{base + nonfiction books + newspapers})
    (38.88,{base + nonfiction books})
    (37.29,{base + fiction books})
    (40.00,{base + books + newspapers})
    (52.85,{base + newspapers})
    (38.68,{base + books})
    (41.55,base)
    (40.20,extended)
};

\addplot[fill=yellow!30, nearly opaque] coordinates {
    (59.99,{Mistral 7B v0.1})
    (60.52,{base (warm)})
    (61.12,{extended (warm)})
    (62.15,{base + translated books})
    (60.95,{base + original books + newspapers})
    (60.83,{base + original books})
    (61.10,{base + nonfiction books + newspapers})
    (60.14,{base + nonfiction books})
    (64.33,{base + fiction books})
    (61.27,{base + books + newspapers})
    (60.85,{base + newspapers})
    (61.24,{base + books})
    (59.66,base)
    (60.85,extended)
};

\draw[dashed, gray] (rel axis cs:0.7525,0) -- (rel axis cs:0.7525,1);

\node[font=\footnotesize] at (axis cs:520,{Mistral 7B v0.1}) {448.01};
\node[font=\footnotesize] at (axis cs:520,{base (warm)}) {480.28};
\node[font=\footnotesize] at (axis cs:520,{extended (warm)}) {482.98};
\node[font=\footnotesize] at (axis cs:520,{base + translated books}) {409.54};
\node[font=\footnotesize] at (axis cs:520,{base + original books + newspapers}) {436.81};
\node[font=\footnotesize] at (axis cs:520,{base + original books}) {427.87};
\node[font=\footnotesize] at (axis cs:520,{base + nonfiction books + newspapers}) {\underline{440.97}};
\node[font=\footnotesize] at (axis cs:520,{base + nonfiction books}) {427.40};
\node[font=\footnotesize] at (axis cs:520,{base + fiction books}) {408.20};
\node[font=\footnotesize] at (axis cs:520,{base + books + newspapers}) {435.64};
\node[font=\footnotesize] at (axis cs:520,{base + newspapers}) {\underline{440.35}};
\node[font=\footnotesize] at (axis cs:520,{base + books}) {427.30};
\node[font=\footnotesize] at (axis cs:520,base) {413.98};
\node[font=\bfseries\footnotesize] at (axis cs:520,extended) {441.83};

\legend{
    Sentiment Analysis,
    Fairness \& Truthfulness,
    Reading Comprehension,
    World Knowledge,
    Commonsense Reasoning,
    Norwegian Language,
    Summarization,
    Translation,
    Variation \& Readability
}
\end{axis}
\end{tikzpicture}
    
    \caption{Total summed scores across all skills averaged by task for each model. Best scores among from-scratch models \underline{underlined}, best overall from-scratch in \textbf{bold}. Dashed line at the \textbf{base} score.}
    \label{fig:total_scores}
\end{figure*}
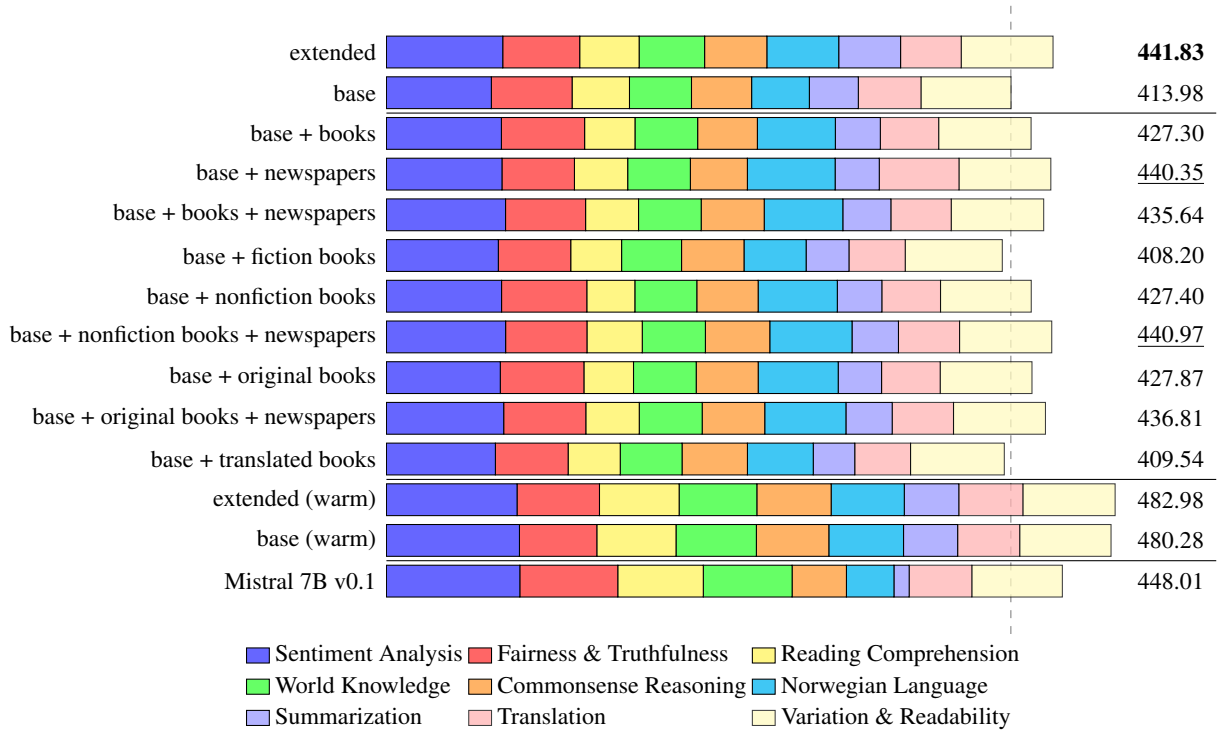

\section{Results} \label{sec:results}
The evaluation of the trained models demonstrated that incorporating publisher-controlled copyright-protected corpora provided a measurable boost in performance across a range of NLP tasks. To illustrate the overall performance differences, Figure~\ref{fig:total_scores} shows the total scores across all skills, averaged by task for each model. Non-aggregated scores for all tasks, prompts, and models are available at the \mimir{} repository.\footnote{\url{https://github.com/mimir-project/mimir-evaluation}}


\subsection{Core Models}


As shown in Table~\ref{tab:ranking} and Figure~\ref{fig:total_scores}, the performance analysis of core models across various tasks reveals distinct strengths for different configurations. The \textbf{base} (warm-started) configuration consistently excels in Sentiment Analysis, World Knowledge, and Norwegian Language. In contrast, the extended (warm-started) configuration leads in Fairness \& Truthfulness, Reading Comprehension, Commonsense Reasoning, Translation, and Variation \& Readability, indicating its robust performance for language-intensive tasks. The \textbf{base} configuration generally lags behind others, scoring the lowest across multiple tasks. Meanwhile, the \textbf{extended} configuration performs well, particularly in Summarization. 
Furthermore, it indicates that we could leverage the existing metadata available at the National Library to tailor subsets of the publisher-controlled copyrighted corpora and build models that excel at specific tasks. However, the difference between the \textbf{base} and \textbf{extended} warm-started models is very small. Further testing is required to assess whether this difference is still statistically significant.\footnote{Detailed scores available in Appendix \ref{app:detailed_scores} Table \ref{tab:detailed_scores}.}

\begin{table}
    \centering
    \resizebox{8cm}{!}{

    \begin{tabular}{lccccccccc}
    \textbf{Model} & \textbf{SA} & \textbf{FT} & \textbf{RC} & \textbf{WK} & \textbf{RC} & \textbf{NL} & \textbf{S} & \textbf{T} & \textbf{VR}\\
    \midrule
    extended  &   3 & 2 & 3 & 3 & 2 & 2 & 1 & 3 & 2  \\
    base       &  4 & 3 & 4 & 4 & 3 & 4 & 3 & 4 & 3  \\
    extended (warm) &   2 & 3 & 1 & 2 & 1 & 1 & 2 & 1 & 1  \\
    base (warm)   &  1 & 1 & 2 & 1 & 1 & 3 & 2 & 2 & 4  \\
    \end{tabular}
    }
    
    \caption{Results for ranking the core models on all tasks by skill via (i) finding the best k-shot configuration for each task and (ii) aggregating metric-wise rankings. SA=Sentiment Analysis. FT=Fairness \& Truthfulness. RC=Reading Comprehension. NL=Norwegian Language. WK=World Knowledge. CR=Commonsense Reasoning. S=Summarization. T=Translation. VR=Variation \& Readability. Lower is better.}
    \label{tab:ranking}
\end{table}

\begin{figure}
\begin{tikzpicture}
\begin{axis}[
    width=5.1cm,
    height=5cm,
    xbar stacked,
    bar width=10pt,
    xlabel={},
    symbolic y coords={
        base + fiction books,
        base + books,
        base + nonfiction books,
        base + original books + newspapers,
        base + newspapers,
        base + nonfiction books + newspapers,
        extended
    },
    ytick=data,
    yticklabel style={font=\scriptsize},
    xticklabel style={font=\scriptsize},
    xmin=-2,
    xmax=9.2,
    reverse legend=false,
    xtick={0},  
    xticklabels={base},  
    axis line style={draw=none},  
    axis x line=top,  
    axis line style={draw=none},
    tick style={draw=none},
    extra x ticks={0},  
    extra x tick style={grid=major, grid style={black}},  
    extra x tick labels={}  
]

\addplot[fill=blue!60] coordinates {
    (0,{base + fiction books})
    (3.22,{base + books})
    (3.24,{base + nonfiction books})
    (6.37,{base + newspapers})
    (6.52,{base + nonfiction books + newspapers})
    (5.51,{base + original books + newspapers})
    (6.73,{extended})
} node[pos=1, anchor=west] {};

\node[anchor=west, font=\scriptsize] at (axis cs:0,{base + fiction books}) {-1.40\%};
\node[anchor=west, font=\scriptsize] at (axis cs:3.22,{base + books}) {3.22\%};
\node[anchor=west, font=\scriptsize] at (axis cs:3.24,{base + nonfiction books}) {3.24\%};
\node[anchor=west, font=\scriptsize] at (axis cs:6.37,{base + newspapers}) {6.37\%};
\node[anchor=west, font=\scriptsize] at (axis cs:6.52,{base + nonfiction books + newspapers}) {6.52\%};
\node[anchor=west, font=\scriptsize] at (axis cs:5.51,{base + original books + newspapers}) {5.51\%};
\node[anchor=west, font=\scriptsize] at (axis cs:6.73,{extended}) {6.73\%};

\addplot[fill=red!60] coordinates {
    (-1.40,{base + fiction books})
    (0,{base + books})
    (0,{base + nonfiction books})
    (0,{base + newspapers})
    (0,{base + nonfiction books + newspapers})
    (0,{base + original books + newspapers})
    (0,{extended})
};

\end{axis}
\end{tikzpicture}
    \caption{Average percentage gains over the performance of the base model. Negative results indicate a decrease in performance over base, positive results a gain.}
    \label{fig:gain}
\end{figure}
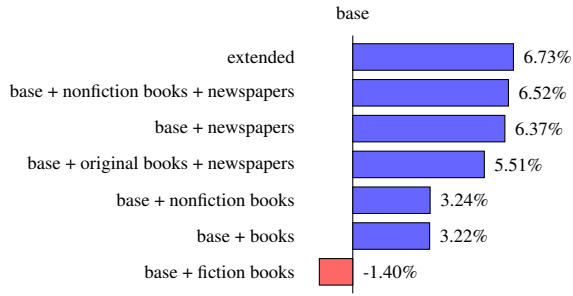


While warm-started models generally outperformed models trained from scratch, there was reduced sensitivity to the presence of copyrighted materials. This suggests that the pre-existing weights, which were primarily trained on English data, diminished the impact of adding high-quality Norwegian copyrighted texts (see also Section \ref{sec:discussion}).

\subsection{Domain-tuned Models}
To further explore the specific effects of different types of training data, we analyzed the gains in performance by focusing on different sub-corpora, such as newspapers, books, and mixed datasets. Figure~\ref{fig:gain} provides an overview of the average percentage gains for models trained on various data configurations compared to the \textbf{base} model.
It shows that the \textbf{extended} model exhibits the highest average gain at 6.73\%, indicating substantial overall improvement. The addition of nonfiction books and newspapers follows with a 6.52\% gain, and the addition of only newspapers shows a 6.37\% improvement. Other configurations, such as adding original books and newspapers or nonfiction books, also demonstrate positive gains of 5.51\% and 3.22\%, respectively. Conversely, the addition of fiction books is the only one to show a negative performance, with a decrease of 1.40\%. 
Interestingly, when decomposed by skill, the addition of fiction books makes the model excel at generating more diverse texts (see Figure \ref{fig:gains} in Appendix \ref{app:gains}). 

\subsection{Instruction-tuned Models}

\begin{figure}
\begin{tikzpicture}
\begin{axis}[
    width=6.5cm,
    height=6cm,
    xbar stacked,
    bar width=10pt,
    xlabel={Scores},
    symbolic y coords={
        Mistral 7B v0.1,
        extended (warm) instruct,
        extended (warm),
        base (warm) instruct,
        base (warm),
        extended instruct,
        extended,
        base instruct,
        base,
    },
    yticklabel style={font=\scriptsize},
    yticklabels={
        extended,
        extended \textit{instruct},
        base,
        base \textit{instruct},
        extended (warm),
        extended (warm) \textit{instruct},
        base (warm),
        base (warm) \textit{instruct},
        Mistral 7B v0.1
    },
    ytick=data,
    legend style={
        at={(0.25,0.0)},
        anchor=north,
        legend columns=3,
        cells={anchor=west},
        font=\tiny,
        draw=none, 
    },
    xmin=0,
    xmax=590,
    legend image code/.code={
        \draw[#1] (0cm,-0.1cm) rectangle (0.3cm,0.1cm);
    },
    reverse legend=false,
    nodes near coords align={left},
    axis line style={draw=none},
    tick style={draw=none},
    xtick=\empty,
    xlabel={}
]



\addplot[fill=blue!60] coordinates {
    (77.09,extended)
    (78.90,{extended instruct})
    (69.54,base)
    (69.45,{base instruct})
    (86.57,{extended (warm)})
    (89.81,{extended (warm) instruct})
    (88.17,{base (warm)})
    (87.83,{base (warm) instruct})
    (88.41,{Mistral 7B v0.1})
};

\addplot[fill=red!60] coordinates {
    (51.01,extended)
    (46.10,{extended instruct})
    (53.51,base)
    (50.59,{base instruct})
    (54.64,{extended (warm)})
    (57.80,{extended (warm) instruct})
    (51.28,{base (warm)})
    (53.70,{base (warm) instruct})
    (64.93,{Mistral 7B v0.1})
};

\addplot[fill=yellow!60] coordinates {
    (39.38,extended)
    (38.68,{extended instruct})
    (38.04,base)
    (36.27,{base instruct})
    (52.79,{extended (warm)})
    (51.69,{extended (warm) instruct})
    (52.48,{base (warm)})
    (50.33,{base (warm) instruct})
    (56.68,{Mistral 7B v0.1})
};

\addplot[fill=green!60] coordinates {
    (43.40,extended)
    (44.32,{extended instruct})
    (41.10,base)
    (41.18,{base instruct})
    (51.51,{extended (warm)})
    (53.09,{extended (warm) instruct})
    (53.27,{base (warm)})
    (54.98,{base (warm) instruct})
    (58.86,{Mistral 7B v0.1})
};

\addplot[fill=orange!60] coordinates {
    (41.26,extended)
    (43.57,{extended instruct})
    (39.85,base)
    (42.06,{base instruct})
    (49.25,{extended (warm)})
    (49.76,{extended (warm) instruct})
    (48.02,{base (warm)})
    (49.42,{base (warm) instruct})
    (36.01,{Mistral 7B v0.1})
};

\addplot[fill=cyan!60] coordinates {
    (47.64,extended)
    (56.46,{extended instruct})
    (38.28,base)
    (53.53,{base instruct})
    (48.48,{extended (warm)})
    (55.91,{extended (warm) instruct})
    (49.51,{base (warm)})
    (59.53,{base (warm) instruct})
    (31.49,{Mistral 7B v0.1})
};

\addplot[fill=blue!30] coordinates {
    (41.00,extended)
    (36.40,{extended instruct})
    (32.45,base)
    (35.14,{base instruct})
    (36.14,{extended (warm)})
    (37.75,{extended (warm) instruct})
    (35.88,{base (warm)})
    (38.36,{base (warm) instruct})
    (10.09,{Mistral 7B v0.1})
};

\addplot[fill=red!30, nearly opaque] coordinates {
    (40.20,extended)
    (54.64,{extended instruct})
    (41.55,base)
    (58.83,{base instruct})
    (42.48,{extended (warm)})
    (47.72,{extended (warm) instruct})
    (41.14,{base (warm)})
    (49.75,{base (warm) instruct})
    (41.55,{Mistral 7B v0.1})
};

\addplot[fill=yellow!30, nearly opaque] coordinates {
    (60.85,extended)
    (59.29,{extended instruct})
    (59.66,base)
    (58.35,{base instruct})
    (61.12,{extended (warm)})
    (60.35,{extended (warm) instruct})
    (60.52,{base (warm)})
    (59.46,{base (warm) instruct})
    (59.99,{Mistral 7B v0.1})
};

\draw[dashed, gray] (rel axis cs:0.702,0) -- (rel axis cs:0.702,1);

\node[font=\scriptsize] at (axis cs:550,extended) {441.83};
\node[font=\scriptsize] at (axis cs:550,{extended instruct}) {458.36};
\node[font=\scriptsize] at (axis cs:550,base) {413.98};
\node[font=\scriptsize] at (axis cs:550,{base instruct}) {445.39};
\node[font=\scriptsize] at (axis cs:550,{extended (warm)}) {\underline{482.98}};
\node[font=\bfseries\scriptsize] at (axis cs:550,{extended (warm) instruct}) {503.89};
\node[font=\scriptsize] at (axis cs:550,{base (warm)}) {480.28};
\node[font=\bfseries\scriptsize] at (axis cs:550,{base (warm) instruct}) {503.36};
\node[font=\scriptsize] at (axis cs:550,{Mistral 7B v0.1}) {448.01};

\legend{
    Sentiment Analysis,
    Fairness \& Truthfulness,
    Reading Comprehension,
    World Knowledge,
    Commonsense Reasoning,
    Norwegian Language,
    Summarization,
    Translation,
    Variation \& Readability
}
\end{axis}
\end{tikzpicture}
    \caption{Total scores (sum) of all averaged scores per skill for the core models and their instruct versions, with original Mistral 7B v0.1 for reference. Dashed line at the \textbf{base} score. Best scores in \textbf{bold}, second best \underline{underlined}.}
    \label{fig:instruct}
\end{figure}
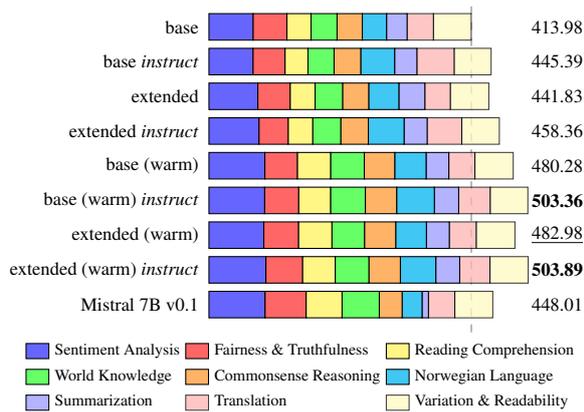

Lastly, as shown in Figure~\ref{fig:instruct}, when the core models are further fine-tuned on data to follow instructions, the gains across models are all consistent, showing that the core advantage lies in the pre-training data, while further training on instructions gives a consistent boost in performance. Instruction tuning also seems to reduce the gap between the \textbf{base} and \textbf{extended} configurations, suggesting that publisher-controlled copyrighted corpora might become less relevant as supervised fine-tuning datasets increase in size in the post-training phases of LLMs. Interestingly, adding Norwegian instruction data on top of the \textbf{extended} model seems enough to improve over the performance of Mistral 7B v0.1.

\section{Discussion}\label{sec:discussion}
Our findings underline the value of copyrighted materials in improving the performance of generative language models, particularly for specialized NLP tasks in Norwegian. The inclusion of these curated publisher-controlled texts provide a substantial advantage in terms of language richness, coherence, and context-specific understanding. 
However, these advantages are significantly less evident in models that are warm-started using weights pre-trained on other languages, primarily English. We see two possible reasons for this:
\begin{enumerate}\itemsep0em 
    \item The \textit{amount} of training data matters more than its quality or licensing status. Warm-started models are effectively trained on more data than the `from-scratch' models, and at some point adding even more data brings diminishing returns (with a given model size).
    \item Publisher-controlled copyrighted Norwegian data is indeed beneficial for LLMs, but the original models used for warm-starting \textit{were presumably already pre-trained on datasets that may share similarities with this data}. Due to the lack of transparency regarding the exact composition of training datasets in models like Mistral, concerns about potential data contamination remain relevant. This overlap could explain why continuous pre-training on similar content did not yield the expected benefits for the warm-started extended models \cite{li2023open, dong2024generalization, xu2024benchmark, samuel2024towards}.
\end{enumerate} 

\subsection{Ethical and Legal Considerations}
The use of copyrighted materials in model training raises significant ethical and legal questions. The observed improvements in model quality must be balanced against the rights of content creators, who have not consented to the use of their work. This highlights the need for guidelines and compensation mechanisms that recognize the value of copyrighted materials in LLM development.

\subsection{Implications for Policy}
The empirical evidence gathered in our research is crucial for informing copyright policy in the digital age. Policymakers can use these findings to establish frameworks that ensure creators are adequately compensated, balancing the needs of LLM innovation with the rights of authors and publishers. This is particularly relevant in light of ongoing lawsuits against major AI companies. 

\section{Conclusion}
Our study represents a pioneering effort to quantify the impact of copyrighted materials on LLMs for Norwegian. Our results indicate that high-quality publisher-controlled copyrighted content significantly enhances model performance, especially for complex NLP tasks. However, these benefits bring forth ethical and legal challenges that must be addressed to ensure a sustainable and fair approach to LLM development. By providing empirical evidence, we aim to contribute to the ongoing discourse on the role of copyright in AI and inform future policies that support both innovation and the rights of content creators.




\section{Future Work}
Future work should focus on testing models at various scales and different pre-trained open weights to better understand how dataset composition affects performance. By experimenting with models of different sizes, we could identify any scaling thresholds where the impact of copyrighted material varies significantly. In retrospect, one notable flaw in the experimental design is the lack of fully traceable and transparent models, such as OLMo \cite{groeneveld-etal-2024-olmo}, which provide detailed documentation of their training data and processes. Without utilizing models with verifiable data provenance, it becomes challenging to accurately assess how specific dataset compositions, including copyrighted or genre-specific materials, influence model behavior and performance for warm-started models. Incorporating traceable models would improve the reproducibility and reliability of findings, ensuring that conclusions drawn about the impact of various text genres are well-founded.

Additionally, the observed effects of fiction on model performance highlight the need to 1) examine how different types of fiction --such as fantasy or historical fiction-- impact tasks like Sentiment Analysis and Commonsense Reasoning, and 2) design new and adequate benchmarks for evaluating the contribution of fiction in Norwegian LLMs for tasks such as creative writing, plot understanding, or descriptive language use. This investigation could clarify the role of fiction in model training and help refine data curation strategies.

Lastly, exploring genre-specific influences more deeply, including essays, technical writing, and narrative nonfiction, may reveal distinct benefits or biases tied to each genre. Analyzing these nuances, even in a diachronic manner, will guide balanced genre representation in datasets and support the development of better performing models.

\section{Distribution}
The \textbf{base} dataset and models were intended to be freely distributed, as all materials included were granted redistribution permissions under different agreements. After we communicated the results of our investigations to the different partners, some right-holders demanded a reinterpretation of the agreements (primarily the Language Technology Use, \textit{Språkteknologiformål}), in the light of the results and this new era of generative AI. This prevented us from sharing publicly the exact models trained within the \mimir{} project, but instead we built a subset of \textbf{base}, which we are calling \textbf{core}, excluding the affected newspapers (around 1B words) and trained models both from scratch and from Mistral 7B v0.1. Their performance is on par with their \textbf{base} counterparts. We are also releasing these models under a permissive license.\footnote{\url{https://huggingface.co/mimir-project/mimir-mistral-7b-core-scratch} and \url{https://huggingface.co/mimir-project/mimir-mistral-7b-core}}

\section*{Acknowledgments}
We extend our sincere gratitude to Hans Eide from Sigma2 for facilitating access to the LUMI supercomputer, enabling the computationally intensive tasks integral to this study. Additionally, we thank Google for providing compute resources via the Tensor Research Cloud program, which significantly supported our model training efforts.

This project would not have been possible without the trust and collaboration of the Ministry of Culture and Equality, which empowered the National Library of Norway to spearhead this endeavor, with the invaluable contributions of the Norwegian University of Science and Technology (NTNU) and the University of Oslo (UiO), whose expertise and insights were instrumental throughout the process. We are grateful for their vision and faith in the potential of this research.

We are also deeply appreciative of Olaus Bergstrøm and the entire legal team at the National Library of Norway for their guidance on the legal dimensions of this research. Their expertise was invaluable in navigating the complexities of copyright law and ensuring compliance with the unique considerations surrounding the materials used in this project.

A special thanks goes to the representatives of the Norwegian rights-holder organizations, who not only agreed to the use of their materials for this project but were steadfast in their support of the initiative. Their cooperation and encouragement have been vital in ensuring the project’s success and advancing research on the intersection of copyright and AI development.

\bibliographystyle{acl_natbib}
\bibliography{anthology_0, custom}

\appendix
\section*{Appendices}
\section{Legal Cases}\label{app:legal_cases}

\begin{itemize}
\small
\itemsep0em
  \item \textbf{Bartz v. Anthropic PBC}, No. 3:24-cv-05417 (N.D. Cal. Aug. 19, 2024)
  \item \textbf{The Ctr. for Investigative Reporting v. OpenAI, Inc.}, No. 1:24-cv-04872 (S.D.N.Y. Jun. 27, 2024)
  \item \textbf{UMG Recordings, Inc. v. Uncharted Labs, LLC}, No. 1:24-cv-04777 (S.D.N.Y. Jun. 24, 2024)
  \item \textbf{UMG Recordings, Inc. v. Suo, Inc.}, No. 1:24-cv-11611 (D. Mass. Jun. 24, 2024)
  \item \textbf{J. L. v. Alphabet, Inc.}, No. 3:23-cv-03440, 2024 WL 3282528 (N.D. Cal. June 6, 2024)
  \item \textbf{In re OpenAI ChatGPT Litigation}, No. 3:23-cv-03223, 2024 WL 2044625 (N.D. Cal. May 7, 2024)
  \item \textbf{Makkai v. Databricks, Inc.}, No. 4:24-cv-02653 (N.D. Cal. May 2, 2024)
  \item \textbf{Dubus v. NVIDIA Corp.}, No. 3:24-cv-02655 (N.D. Cal. May 2, 2024)
  \item \textbf{Daily News LP v. Microsoft Corp.}, No. 1:24-cv-03285 (S.D.N.Y. Apr. 30, 2024)
  \item \textbf{Zhang v. Google LLC}, No. 3:24-cv-02531 (N.D. Cal. Apr. 26, 2024)
  \item \textbf{Nazemian v. NVIDIA Corp.}, No. 4:24-cv-01454 (N.D. Cal. Mar. 8, 2024)
  \item \textbf{The Intercept Media, Inc. v. OpenAI, Inc.}, No. 1:24-cv-01515 (S.D.N.Y. Feb. 28, 2024)
  \item \textbf{Raw Story Media, Inc. v. OpenAI Inc.}, No. 1:24-cv-01514 (S.D.N.Y. Feb. 28, 2024)
  \item \textbf{Tremblay v. OpenAI, Inc.}, No. 3:23-cv-03233, 2024 WL 557720 (N.D. Cal. Feb. 12, 2024)
  \item \textbf{Universal Music Group v. Anthropic} (February 2024)
  \item \textbf{The New York Times v. OpenAI \& Microsoft} (December 2023)
  \item \textbf{Kadrey v. Meta Platforms, Inc.}, No. 3:23-cv-03417, 2023 WL 10673221 (N.D. Cal. Dec. 1, 2023)
  \item \textbf{Alter v. OpenAI Inc.}, No. 1:23-cv-10211 (S.D.N.Y. Nov. 21, 2023)
  \item \textbf{Andersen v. Stability AI Ltd.}, No. 3:23-cv-00201, 2023 WL 7132064 (N.D. Cal. Oct. 30, 2023)
  \item \textbf{Concord Music Group, Inc. v. Anthropic PBC}, No. 3:23-cv-01092 (M.D. Tenn. Oct. 18, 2023)
  \item \textbf{Huckabee v. Meta Platforms, Inc.}, No. 3:23-cv-06663 (N.D. Cal. Oct. 17, 2023)
  \item \textbf{Authors Guild v. OpenAI Inc.}, No. 1:23-cv-08292 (S.D.N.Y. Sept. 18, 2023)
  \item \textbf{Silverman v. OpenAI Inc.}, No. 3:23-cv-03416 (N.D. Cal. July 7, 2023).
  \item \textbf{Thaler v. Perlmutter}, 687 F. Supp. 3d 140 (D.D.C. 2023)
  \item \textbf{Doe 1 v. Github, Inc.}, No. 4:22-cv-06823, 2023 WL 3449131 (N.D. Cal. May 11, 2023)
  \item \textbf{Getty Images (US), Inc. v. Stability AI, Inc.}, No. 1:23-cv-00135 (D. Del. Feb. 3, 2023)
\end{itemize}

\vfill 
\newpage 
\onecolumn

\section{Sampling}\label{app:sampling}
We built three custom perplexity models for specific Norwegian domains that proved too divergent from Wikipedia: books, newspapers, and government documents. These perplexity models were used to score each document in the datasets. Based on their perplexity scores, the documents were further divided into three segments corresponding to their quartile distribution. Documents with scores below the first quartile were classified as ``good'', those between $Q_1$ and $Q_3$ as ``medium'', and those above $Q_3$ were considered ``bad''. The documents in each segment were randomized. While the intention was to train all models on progressively better data, starting from ``bad'' segment, then ``medium'' and finally the ``good'' segment, we never got around to test whether this approach would result in better performing models.

Moreover, from the clean and deduplicated sources, we sub-sampled each non-Norwegian language at an specific sampling ratio until achieving the proportion of documents shown in Figure \ref{tab:language_ratios}. Pseudo-code for the algorithm used to subsample is shown in Algorithm \ref{algo:subsample}.\footnote{\url{https://huggingface.co/mimir-project/mimir-perplexity}} We also discovered that a good amount of documents were misclassified by the fastText language identifier \cite{joulin2016bag}.

\begin{table}[h!]
\centering
\begin{tabular}{lcc}
\textbf{Language} & \textbf{Sampling ratio} & \textbf{Final ratio} \\
\hline
Bokmål & 100.00\% & 35.74\% \\
Danish & 43.00\% & 8.01\% \\
English & 81.00\% & 4.53\% \\
Icelandic & 100.00\% & 1.31\% \\
Nynorsk & 100.00\% & 2.02\% \\
Swedish & 15.40\% & 4.46\% \\
\hline
Code & 62.00\% & 4.53\% \\
\end{tabular}
\caption{Percentage of documents kept from the clean and deduplicated sources and the final proportion of documents in each language present in the final dataset. Code was considered its own language when sampling.}
\label{tab:language_ratios}
\end{table}

\begin{algorithm*}\
     \caption{Sub-sampling Dataset Based on Perplexity Distribution}\label{algo:subsample}
     \begin{algorithmic}[1]
     \State \textbf{Input:} Dataset $D$ with perplexity distribution, target sampling ratio $R$
     \State \textbf{Output:} Sub-sampled dataset $D'$
   
     \Procedure{Subsample}{$D, R$}
         \State Compute the quartile values $q_1$ and $q_3$ from the perplexity distribution of $D$
         \State Define an initial Gaussian PDF with mean $\mu = (q_1 + q_3)/2$ and standard deviation $\sigma$ such that $q_1$ and $q_3$ align with the corresponding positions in the Gaussian curve
         \State Compute the histogram $H$ of perplexity values from $D$
         \State Combine $H$ with the Gaussian weights to estimate the initial sampling ratio $R_0$
         \State Compute the normalization factor $N$ such that $R_0 \times N = R$
       
         \While{Error in central quartile probabilities exceeds tolerance}
             \State Adjust the parameters $\mu$ and $\sigma$ of the Gaussian curve to minimize the error in the desired probabilities within the central quartiles $[q_1, q_3]$
             \State Update the normalization factor $N$ to match the target ratio $R$
         \EndWhile
       
         \For{each sample $s$ in $D$}
             \State Compute the perplexity $p_s$ of sample $s$
             \State Estimate the probability $P(s)$ of retaining sample $s$ based on the normalized Gaussian PDF
             \If{$P(s) \geq$ random threshold}
                 \State Retain $s$ in the sub-sampled dataset $D'$
             \EndIf
         \EndFor
     \EndProcedure
   
     \State \textbf{return} $D'$
     \end{algorithmic}
\end{algorithm*}

\section{Sources}\label{app:sources}

\begin{table}[h!]
\centering
\begin{tabular}{lrrrr}
\textbf{Source} &  \textbf{Raw} & \textbf{Clean} & \textbf{extended} &     \textbf{base} \\
\midrule
Books                & 3.7B   & 2.5B   & 1.9B   & 1.9B   \\
CulturaX             & 52.7B  & 52.1B  & 21.8B  & 16.9B  \\
Digimanus            & 9.6M   & 4.6M   & 3.4M   & 3.3M   \\
Evaluerings- rapport & 76.7M  & 68.6M  & 61.2M  & 61.5M  \\
HPTL v1.2            & 35.5B  & 34.1B  & 14.9B  & 11.3B  \\
LovData              & 57.1M  & 57.1M  & 53.7M  & 54.8M  \\
Målfrid              & 7.5B   & 1.9B   & 1.7B   & 1.7B   \\
Newspapers            & 4.6B   & 3.6B   & 3.2B   & 3.3B   \\
Parlamint            & 170.3M & 84.4M  & 83.4M  & 83.3M  \\
PG19                 & 2.0B   & 1.9B   & 1.4B   & 428.6M \\
StarCoder            & 19.7B  & 9.8B   & 7.1B   & 3.4B   \\
Wikipedia            & 4.0B   & 3.9B   & 2.8B   & 996.2M \\
\midrule
Books (restricted)   & 21.7B  & 20.0B  & 18.1B  & 0   \\
Newspapers (restricted) & 14.3B  & 9.8B   & 9.1B   & 0   \\
\midrule
\textbf{Total} & \textbf{166.1B} & \textbf{139.8B} & \textbf{82.1B} & \textbf{40.1B} \\
\end{tabular}
\caption{Number of words (comma separated) per source at the start of the data pipeline (raw count), after cleaning, and in the \textbf{extended} and \textbf{base} datasets.}
\label{tab:pipeline_sources}
\end{table}

\section{Hyperparameters}\label{app:hyperparams}

\begin{table}[hp!]
  \footnotesize
  \centering
  \setlength\tabcolsep{3.9pt}
  \begin{tabular}{llll}
     \textbf{Hyperparameter} & \textbf{Core Models} & \textbf{Domain-Tuned Models} & \textbf{Instruction-tuned Models} \\ 
    \midrule
     Model size & 7B & 7B & 7B \\
    Hidden layers & 32 & 32 & 32 \\
    Attention heads & 32 & 32 & 32 \\
    Hidden size & 4096 & 4096 & 4096 \\
    Intermediate size & 14336 & 14336 & 14336 \\
    Max position embeddings & 2048 & 2048 & 2048 \\
    Key-value heads & 8 & 8 & 8 \\
    Sliding window & 4096 & 4096 & 4096 \\
    Precision & \texttt{bfloat16} & \texttt{bfloat16} & \texttt{bfloat16} \\
    Optimizer & AdamW & AdamW & AdamW \\
    Optimizer parameters & \tiny{$\beta_1=0.9, \beta_2=0.95, \epsilon=10^{-8}$} & \tiny{$\beta_1=0.9, \beta_2=0.95, \epsilon=10^{-8}$} & \tiny{$\beta_1=0.9, \beta_2=0.95, \epsilon=10^{-8}$} \\
    Global batch size & 4M ($2048 \times 2048$) tokens & 4M ($2048 \times 2048$) tokens & 512 seqs\\
    Initial/final learning rate & $3.0 \times 10^{-4}$ / $3.0 \times 10^{-5}$ & $3.0 \times 10^{-5}$ / $3.0 \times 10^{-6}$  & $3.0 \times 10^{-6}$ / $3.0 \times 10^{-7}$\\
    Vocabulary size & 32768 & 32768 & 32768\\
    Training steps & 64k & 10k & 4 epochs\\
    Dropout & 0 & 0 & 0\\
    Warm-up steps & 2000 & 200 & 20\\
    Weight decay & 0.1 & 0.1 & 0.1\\
    Checkpoints & Every 1000 steps & Every 1000 steps & Every 1 epoch\\
    Shuffle & Shuffle after each epoch & Shuffle after each epoch & Shuffle after each epoch\\
  \end{tabular}
  \caption{Hyperparameters for the \mimir{} model set.}
  \label{tab:hyperparams}
\end{table}

\newpage
\section{Percentage Gains}\label{app:gains}

\begin{figure*}[h!]
\begin{tikzpicture}
\begin{groupplot}[
    group style={
        group size=3 by 3,
        vertical sep=1.5cm
    },
    title style = {font=\small},
    width=4.9cm,
    height=6cm,
    xbar stacked,
    symbolic y coords={
        base + translated books,
        base + original books + newspapers,
        base + original books,
        base + nonfiction books + newspapers,
        base + nonfiction books,
        base + fiction books,
        base + books + newspapers,
        base + newspapers,
        base + books,
        extended
    },
    ytick=data,
    yticklabel style={font=\scriptsize},
    xticklabel style={font=\scriptsize},
    reverse legend=false,
    axis line style={draw=none},
    axis x line=bottom,
    tick style={draw=none},
    extra x ticks={0},
    extra x tick style={grid=major, grid style={black}},
    extra x tick labels={}
]

\nextgroupplot[xtick={0}, xticklabels={base}, xmin=0, xmax=19, title={Sentiment Analysis}]
\addplot[fill=blue!60] coordinates {
    (3.85,{base + translated books})
    (6.64,{base + fiction books})
    (9.58,{base + books})
    (9.72,{base + nonfiction books})
    (10.11,{base + newspapers})
    (13.59,{base + nonfiction books + newspapers})
    (11.77,{base + original books + newspapers})
    (13.52,{base + books + newspapers})
    (8.52,{base + original books})
    (10.86,{extended})
};
\node[anchor=west, font=\scriptsize] at (axis cs:3.85,{base + translated books}) {3.85\%};
\node[anchor=west, font=\scriptsize] at (axis cs:6.64,{base + fiction books}) {6.64\%};
\node[anchor=west, font=\scriptsize] at (axis cs:9.58,{base + books}) {9.58\%};
\node[anchor=west, font=\scriptsize] at (axis cs:9.72,{base + nonfiction books}) {9.72\%};
\node[anchor=west, font=\scriptsize] at (axis cs:10.11,{base + newspapers}) {10.11\%};
\node[anchor=west, font=\scriptsize] at (axis cs:13.59,{base + nonfiction books + newspapers}) {13.59\%};
\node[anchor=west, font=\scriptsize] at (axis cs:11.77,{base + original books + newspapers}) {11.77\%};
\node[anchor=west, font=\scriptsize] at (axis cs:13.52,{base + books + newspapers}) {13.52\%};
\node[anchor=west, font=\scriptsize] at (axis cs:8.52,{base + original books}) {8.52\%};
\node[anchor=west, font=\scriptsize] at (axis cs:10.86,{extended}) {10.86\%};

\nextgroupplot[yticklabels={}, xtick={0}, xticklabels={base}, xmin=-11, xmax=12, title={Fairness \& Truthfulness}]
\addplot[fill=blue!60] coordinates {
    (0,{base + translated books})
    (0,{base + fiction books})
    (3.06,{base + books})
    (5.60,{base + nonfiction books})
    (0,{base + newspapers})
    (0.74,{base + nonfiction books + newspapers})
    (1.71,{base + original books + newspapers})
    (0,{base + books + newspapers})
    (3.59,{base + original books})
    (0,{extended})
};
\addplot[fill=red!60] coordinates {
    (-9.90,{base + translated books})
    (-10.33,{base + fiction books})
    (0,{base + books})
    (0,{base + nonfiction books})
    (-10.23,{base + newspapers})
    (0,{base + nonfiction books + newspapers})
    (0,{base + original books + newspapers})
    (-0.84,{base + books + newspapers})
    (0,{base + original books})
    (-4.67,{extended})
};
\node[anchor=west, font=\scriptsize] at (axis cs:3.06,{base + books}) {3.06\%};
\node[anchor=west, font=\scriptsize] at (axis cs:5.60,{base + nonfiction books}) {5.60\%};
\node[anchor=west, font=\scriptsize] at (axis cs:0.74,{base + nonfiction books + newspapers}) {0.74\%};
\node[anchor=west, font=\scriptsize] at (axis cs:1.71,{base + original books + newspapers}) {1.71\%};
\node[anchor=west, font=\scriptsize] at (axis cs:0,{base + translated books}) {-9.90\%};
\node[anchor=west, font=\scriptsize] at (axis cs:0,{base + fiction books}) {-10.33\%};
\node[anchor=west, font=\scriptsize] at (axis cs:0,{base + newspapers}) {-10.23\%};
\node[anchor=west, font=\scriptsize] at (axis cs:0,{base + books + newspapers}) {-0.84\%};
\node[anchor=west, font=\scriptsize] at (axis cs:3.59,{base + original books}) {3.59\%};
\node[anchor=west, font=\scriptsize] at (axis cs:0,{extended}) {-4.67\%};

\nextgroupplot[yticklabels={}, xtick={0}, xticklabels={base}, xmin=-17, xmax=11, title={Reading Comprehension}]
\addplot[fill=blue!60] coordinates {
    (0,{base + translated books})
    (0,{base + fiction books})
    (0,{base + books})
    (0,{base + nonfiction books})
    (0,{base + newspapers})
    (0,{base + nonfiction books + newspapers})
    (0,{base + original books + newspapers})
    (0,{base + books + newspapers})
    (0,{base + original books})
    (3.52,{extended})
};
\addplot[fill=red!60] coordinates {
    (-9.20,{base + translated books})
    (-11.37,{base + fiction books})
    (-12.24,{base + books})
    (-16.30,{base + nonfiction books})
    (-7.22,{base + newspapers})
    (-3.63,{base + nonfiction books + newspapers})
    (-6.89,{base + original books + newspapers})
    (-7.75,{base + books + newspapers})
    (-13.59,{base + original books})
    (0,{extended})
};
\node[anchor=west, font=\scriptsize] at (axis cs:0,{base + translated books}) {-9.20\%};
\node[anchor=west, font=\scriptsize] at (axis cs:0,{base + fiction books}) {-11.37\%};
\node[anchor=west, font=\scriptsize] at (axis cs:0,{base + books}) {-12.24\%};
\node[anchor=west, font=\scriptsize] at (axis cs:0,{base + nonfiction books}) {-16.30\%};
\node[anchor=west, font=\scriptsize] at (axis cs:0,{base + newspapers}) {-7.22\%};
\node[anchor=west, font=\scriptsize] at (axis cs:0,{base + nonfiction books + newspapers}) {-3.63\%};
\node[anchor=west, font=\scriptsize] at (axis cs:0,{base + original books + newspapers}) {-6.89\%};
\node[anchor=west, font=\scriptsize] at (axis cs:0,{base + books + newspapers}) {-7.75\%};
\node[anchor=west, font=\scriptsize] at (axis cs:0,{base + original books}) {-13.59\%};
\node[anchor=west, font=\scriptsize] at (axis cs:3.52,{extended}) {3.52\%};

\nextgroupplot[xtick={0}, xticklabels={base}, xmin=-4, xmax=10, title={World Knowledge}]
\addplot[fill=blue!60] coordinates {
    (0,{base + translated books})
    (0,{base + fiction books})
    (1.08,{base + books})
    (0,{base + nonfiction books})
    (1.08,{base + newspapers})
    (1.82,{base + nonfiction books + newspapers})
    (1.48,{base + original books + newspapers})
    (1.05,{base + books + newspapers})
    (1.13,{base + original books})
    (5.60,{extended})
};
\addplot[fill=red!60] coordinates {
    (-0.31,{base + translated books})
    (-3.32,{base + fiction books})
    (0,{base + books})
    (-0.11,{base + nonfiction books})
    (0,{base + newspapers})
    (0,{base + nonfiction books + newspapers})
    (0,{base + original books + newspapers})
    (0,{base + books + newspapers})
    (0,{base + original books})
    (0,{extended})
};
\node[anchor=west, font=\scriptsize] at (axis cs:1.08,{base + books}) {1.08\%};
\node[anchor=west, font=\scriptsize] at (axis cs:1.08,{base + newspapers}) {1.08\%};
\node[anchor=west, font=\scriptsize] at (axis cs:1.82,{base + nonfiction books + newspapers}) {1.82\%};
\node[anchor=west, font=\scriptsize] at (axis cs:1.48,{base + original books + newspapers}) {1.48\%};
\node[anchor=west, font=\scriptsize] at (axis cs:1.05,{base + books + newspapers}) {1.05\%};
\node[anchor=west, font=\scriptsize] at (axis cs:1.13,{base + original books}) {1.13\%};
\node[anchor=west, font=\scriptsize] at (axis cs:5.60,{extended}) {5.60\%};
\node[anchor=west, font=\scriptsize] at (axis cs:0,{base + nonfiction books}) {-0.11\%};
\node[anchor=west, font=\scriptsize] at (axis cs:0,{base + translated books}) {-0.31\%};
\node[anchor=west, font=\scriptsize] at (axis cs:0,{base + fiction books}) {-3.32\%};

\nextgroupplot[yticklabels={}, xtick={0}, xticklabels={base}, xmin=-6, xmax=14, title={Commonsense Reasoning}]
\addplot[fill=blue!60] coordinates {
    (8.72,{base + translated books})
    (3.86,{base + fiction books})
    (0,{base + books})
    (2.04,{base + nonfiction books})
    (0,{base + newspapers})
    (7.09,{base + nonfiction books + newspapers})
    (4.04,{base + original books + newspapers})
    (4.81,{base + books + newspapers})
    (3.08,{base + original books})
    (3.55,{extended})
};
\addplot[fill=red!60] coordinates {
    (0,{base + translated books})
    (0,{base + fiction books})
    (-0.65,{base + books})
    (0,{base + nonfiction books})
    (-5.18,{base + newspapers})
    (0,{base + nonfiction books + newspapers})
    (0,{base + original books + newspapers})
    (0,{base + books + newspapers})
    (0,{base + original books})
    (0,{extended})
};
\node[anchor=west, font=\scriptsize] at (axis cs:8.72,{base + translated books}) {8.72\%};
\node[anchor=west, font=\scriptsize] at (axis cs:3.86,{base + fiction books}) {3.86\%};
\node[anchor=west, font=\scriptsize] at (axis cs:2.04,{base + nonfiction books}) {2.04\%};
\node[anchor=west, font=\scriptsize] at (axis cs:7.09,{base + nonfiction books + newspapers}) {7.09\%};
\node[anchor=west, font=\scriptsize] at (axis cs:4.04,{base + original books + newspapers}) {4.04\%};
\node[anchor=west, font=\scriptsize] at (axis cs:4.81,{base + books + newspapers}) {4.81\%};
\node[anchor=west, font=\scriptsize] at (axis cs:3.08,{base + original books}) {3.08\%};
\node[anchor=west, font=\scriptsize] at (axis cs:3.55,{extended}) {3.55\%};
\node[anchor=west, font=\scriptsize] at (axis cs:0,{base + books}) {-0.65\%};
\node[anchor=west, font=\scriptsize] at (axis cs:0,{base + newspapers}) {-5.18\%};

\nextgroupplot[yticklabels={}, xtick={0}, xticklabels={base}, xmin=0, xmax=72, title={Norwegian Language}]
\addplot[fill=blue!60] coordinates {
    (13.98,{base + translated books})
    (7.56,{base + fiction books})
    (35.03,{base + books})
    (37.10,{base + nonfiction books})
    (51.92,{base + newspapers})
    (42.37,{base + nonfiction books + newspapers})
    (40.57,{base + original books + newspapers})
    (36.13,{base + books + newspapers})
    (38.55,{base + original books})
    (24.43,{extended})
};
\node[anchor=west, font=\scriptsize] at (axis cs:13.98,{base + translated books}) {13.98\%};
\node[anchor=west, font=\scriptsize] at (axis cs:7.56,{base + fiction books}) {7.56\%};
\node[anchor=west, font=\scriptsize] at (axis cs:35.03,{base + books}) {35.03\%};
\node[anchor=west, font=\scriptsize] at (axis cs:37.10,{base + nonfiction books}) {37.10\%};
\node[anchor=west, font=\scriptsize] at (axis cs:51.92,{base + newspapers}) {51.92\%};
\node[anchor=west, font=\scriptsize] at (axis cs:42.37,{base + nonfiction books + newspapers}) {42.37\%};
\node[anchor=west, font=\scriptsize] at (axis cs:40.57,{base + original books + newspapers}) {40.57\%};
\node[anchor=west, font=\scriptsize] at (axis cs:36.13,{base + books + newspapers}) {36.13\%};
\node[anchor=west, font=\scriptsize] at (axis cs:38.55,{base + original books}) {38.55\%};
\node[anchor=west, font=\scriptsize] at (axis cs:24.43,{extended}) {24.43\%};

\nextgroupplot[xtick={0}, xticklabels={base}, xmin=-20, xmax=45, title={Summarization}]
\addplot[fill=blue!60] coordinates {
    (0,{base + translated books})
    (0,{base + fiction books})
    (0,{base + books})
    (0,{base + nonfiction books})
    (0,{base + newspapers})
    (0,{base + nonfiction books + newspapers})
    (0,{base + original books + newspapers})
    (0,{base + books + newspapers})
    (0,{base + original books})
    (26.37,{extended})
};
\addplot[fill=red!60] coordinates {
    (-15.21,{base + translated books})
    (-12.43,{base + fiction books})
    (-8.13,{base + books})
    (-9.01,{base + nonfiction books})
    (-9.81,{base + newspapers})
    (-5.13,{base + nonfiction books + newspapers})
    (-5.47,{base + original books + newspapers})
    (-1.80,{base + books + newspapers})
    (-11.43,{base + original books})
    (0,{extended})
};
\node[anchor=west, font=\scriptsize] at (axis cs:26.37,{extended}) {26.37\%};
\node[anchor=west, font=\scriptsize] at (axis cs:0,{base + translated books}) {-15.21\%};
\node[anchor=west, font=\scriptsize] at (axis cs:0,{base + fiction books}) {-12.43\%};
\node[anchor=west, font=\scriptsize] at (axis cs:0,{base + books}) {-8.13\%};
\node[anchor=west, font=\scriptsize] at (axis cs:0,{base + nonfiction books}) {-9.01\%};
\node[anchor=west, font=\scriptsize] at (axis cs:0,{base + newspapers}) {-9.81\%};
\node[anchor=west, font=\scriptsize] at (axis cs:0,{base + nonfiction books + newspapers}) {-5.13\%};
\node[anchor=west, font=\scriptsize] at (axis cs:0,{base + original books + newspapers}) {-5.47\%};
\node[anchor=west, font=\scriptsize] at (axis cs:0,{base + books + newspapers}) {-1.80\%};
\node[anchor=west, font=\scriptsize] at (axis cs:0,{base + original books}) {-11.43\%};

\nextgroupplot[yticklabels={}, xtick={0}, xticklabels={base}, xmin=-12, xmax=43, title={Translation}]
\addplot[fill=blue!60] coordinates {
    (0,{base + translated books})
    (0,{base + fiction books})
    (0,{base + books})
    (0,{base + nonfiction books})
    (27.20,{base + newspapers})
    (0,{base + nonfiction books + newspapers})
    (0,{base + original books + newspapers})
    (0,{base + books + newspapers})
    (0,{base + original books})
    (0,{extended})
};
\addplot[fill=red!60] coordinates {
    (-11.01,{base + translated books})
    (-10.24,{base + fiction books})
    (-6.90,{base + books})
    (-6.42,{base + nonfiction books})
    (0,{base + newspapers})
    (-2.52,{base + nonfiction books + newspapers})
    (-2.18,{base + original books + newspapers})
    (-3.72,{base + books + newspapers})
    (-6.47,{base + original books})
    (-3.25,{extended})
};
\node[anchor=west, font=\scriptsize] at (axis cs:27.20,{base + newspapers}) {27.20\%};
\node[anchor=west, font=\scriptsize] at (axis cs:0,{base + translated books}) {-11.01\%};
\node[anchor=west, font=\scriptsize] at (axis cs:0,{base + fiction books}) {-10.24\%};
\node[anchor=west, font=\scriptsize] at (axis cs:0,{base + books}) {-6.90\%};
\node[anchor=west, font=\scriptsize] at (axis cs:0,{base + nonfiction books}) {-6.42\%};
\node[anchor=west, font=\scriptsize] at (axis cs:0,{base + nonfiction books + newspapers}) {-2.52\%};
\node[anchor=west, font=\scriptsize] at (axis cs:0,{base + original books + newspapers}) {-2.18\%};
\node[anchor=west, font=\scriptsize] at (axis cs:0,{base + books + newspapers}) {-3.72\%};
\node[anchor=west, font=\scriptsize] at (axis cs:0,{base + original books}) {-6.47\%};
\node[anchor=west, font=\scriptsize] at (axis cs:0,{extended}) {-3.25\%};

\nextgroupplot[yticklabels={}, xtick={0}, xticklabels={base}, xmin=0, xmax=11, title={Variation \& Readability}]
\addplot[fill=blue!60] coordinates {
    (4.16,{base + translated books})
    (7.83,{base + fiction books})
    (2.65,{base + books})
    (0.80,{base + nonfiction books})
    (1.98,{base + newspapers})
    (2.41,{base + nonfiction books + newspapers})
    (2.15,{base + original books + newspapers})
    (2.68,{base + books + newspapers})
    (1.96,{base + original books})
    (1.99,{extended})
};
\node[anchor=west, font=\scriptsize] at (axis cs:4.16,{base + translated books}) {4.16\%};
\node[anchor=west, font=\scriptsize] at (axis cs:7.83,{base + fiction books}) {7.83\%};
\node[anchor=west, font=\scriptsize] at (axis cs:2.65,{base + books}) {2.65\%};
\node[anchor=west, font=\scriptsize] at (axis cs:0.80,{base + nonfiction books}) {0.80\%};
\node[anchor=west, font=\scriptsize] at (axis cs:1.98,{base + newspapers}) {1.98\%};
\node[anchor=west, font=\scriptsize] at (axis cs:2.41,{base + nonfiction books + newspapers}) {2.41\%};
\node[anchor=west, font=\scriptsize] at (axis cs:2.15,{base + original books + newspapers}) {2.15\%};
\node[anchor=west, font=\scriptsize] at (axis cs:2.68,{base + books + newspapers}) {2.68\%};
\node[anchor=west, font=\scriptsize] at (axis cs:1.96,{base + original books}) {1.96\%};
\node[anchor=west, font=\scriptsize] at (axis cs:1.99,{extended}) {1.99\%};

\end{groupplot}
\end{tikzpicture}
\caption{Percentage gains over the performance of the \textbf{base} model per skill.}
\label{fig:gains}
\end{figure*}
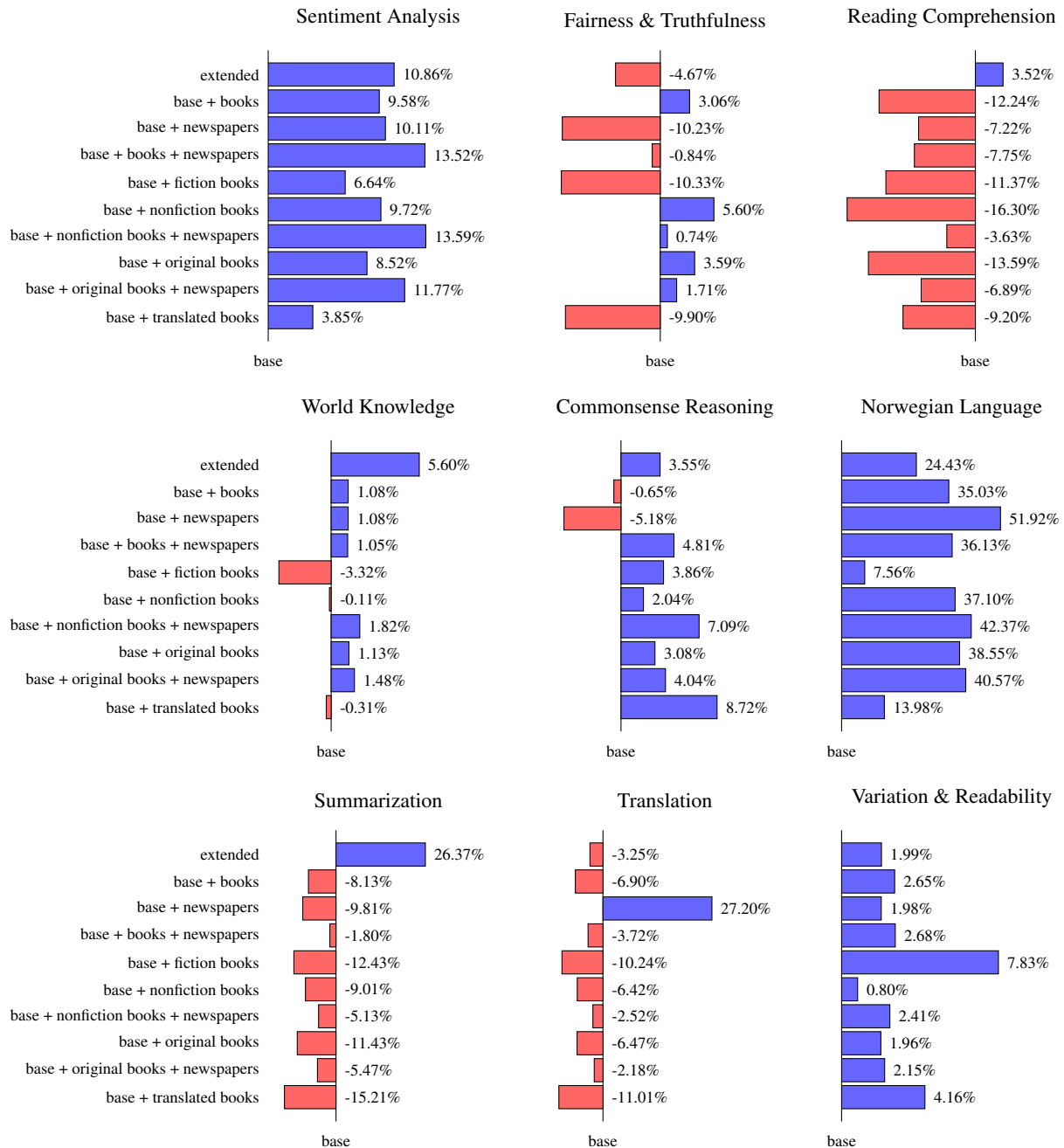

Figure \ref{fig:gains} illustrates the percentage gains of each domain-tuned model with respect to the performance of the \textbf{base} model, per higher level skill. Training on different materials shows distinct trade-offs: newspaper data excels at Translation (27.20\% gain) and Norwegian Language (51.92\%), while fiction books improve Variation \& Readability (7.83\%). Combining books and newspapers often yields balanced improvements, though most configurations struggle with Reading Comprehension and Translation. The \textbf{extended} configuration, which supplements books and newspapers with Internet data, shows strong all-around performance, particularly in Summarization (26.37\%) and World Knowledge (5.60\%).

\newpage
\section{Evaluation Scores}\label{app:detailed_scores}

\begin{table*}[h!]
\centering
\resizebox{\textwidth}{!}{%
\begin{tabular}{lcccccccccc}
\textbf{Model} & \textbf{SA} & \textbf{FT} & \textbf{RC} & \textbf{WK} & \textbf{CR} & \textbf{NL} & \textbf{S} & \textbf{T} & \textbf{VR} & \textbf{Score} \\ \hline
\rowcolor{lightgray!50} \multicolumn{11}{c}{Core Models}\\ \hline
base 
  & 69.54         & 53.51         & 38.04         & 41.10         & 39.85         & 38.28         & 32.45         & 41.55         & 59.66         & 413.98         \\
extended 
  & 77.09         & 51.01         & 39.38         & 43.40         & 41.26         & 47.64         & \textbf{\underline{41.00}} & 40.20         & 60.85         & 441.83         \\
base (warm) 
  & \underline{88.17} & 51.28         & 52.48         & \underline{53.27} & 48.02         & \underline{49.51} & 35.88         & 41.14         & 60.52         & 480.28         \\
extended (warm) 
  & 86.57         & \underline{54.64} & \underline{52.79} & 51.51         & \underline{49.25} & 48.48         & 36.14         & \underline{42.48} & \underline{61.12} & \underline{482.98} \\
\hline
\rowcolor{lightgray!50} \multicolumn{11}{c}{Domain Tuned Models}\\ \hline
base + books 
  & 76.20         & 55.15         & 33.38         & 41.54         & 39.59         & 51.69         & 29.81         & 38.68         & 61.24         & 427.30         \\
base + newspapers 
  & 76.57         & 48.04         & 35.29         & 41.55         & 37.79         & \underline{58.16} & 29.26         & \underline{52.85} & 60.85         & 440.35         \\
base + books + newspapers 
  & 78.94         & 53.06         & 35.09         & 41.53         & 41.77         & 52.11         & \underline{31.86} & 40.00         & 61.27         & 435.64         \\
base + fiction books 
  & 74.16         & 47.99         & 33.71         & 39.74         & 41.39         & 41.18         & 28.41         & 37.29         & \textbf{\underline{64.33}} & 408.20         \\
base + nonfiction books 
  & 76.30         & \underline{56.51} & 31.84         & 41.06         & 40.66         & 52.48         & 29.52         & 38.88         & 60.14         & 427.40         \\
base + nonfiction books + newspapers 
  & \underline{78.99} & 53.91         & \underline{36.66} & \underline{41.85} & 42.68         & 54.50         & 30.78         & 40.50         & 61.10         & \underline{440.97} \\
base + original books 
  & 75.46         & 55.43         & 32.87         & 41.56         & 41.08         & 53.04         & 28.74         & 38.86         & 60.83         & 427.87         \\
base + original books + newspapers 
  & 77.72         & 54.43         & 35.42         & 41.71         & 41.46         & 53.81         & 30.67         & 40.64         & 60.95         & 436.81         \\
base + translated books 
  & 72.22         & 48.21         & 34.54         & 40.97         & \underline{43.33} & 43.63         & 27.51         & 36.97         & 62.15         & 409.54         \\
\hline
\rowcolor{lightgray!50} \multicolumn{11}{c}{Instruction Fine Tuned Models}\\ \hline
base (warm) \textit{instruct} 
  & 87.83         & 53.70         & 50.33         & \underline{54.98} & 49.42         & \textbf{\underline{59.53}} & \underline{38.36} & 49.75         & 59.46         & 503.36         \\
extended (warm) \textit{instruct} 
  & \textbf{\underline{89.81}} & \underline{57.80} & \underline{51.69} & 53.09         & \textbf{\underline{49.76}} & 55.91         & 37.75         & 47.72         & \underline{60.35} & \textbf{\underline{503.89}} \\
base \textit{instruct} 
  & 69.45         & 50.59         & 36.27         & 41.18         & 42.06         & 53.53         & 35.14         & \textbf{\underline{58.83}} & 58.35         & 445.39         \\
extended \textit{instruct} 
  & 78.90         & 46.10         & 38.68         & 44.32         & 43.57         & 56.46         & 36.40         & 54.64         & 59.29         & 458.36         \\
\hline
Mistral 7B v0.1 
  & 88.41 & \textbf{64.93} & \textbf{56.68} & \textbf{58.86} & 36.01 & 31.49 & 10.09 & 41.55 & 59.99 & 448.01 \\

\end{tabular}
}
\caption{Detailed scores across all skills for each model configuration. Abbreviations: SA = Sentiment Analysis, FT = Fairness \& Truthfulness, RC = Reading Comprehension, WK = World Knowledge, CR = Commonsense Reasoning, NL = Norwegian Language, S = Summarization, T = Translation, VR = Variation \& Readability. Best overall scores per skill in \textbf{bold}. Best score per skill and model group \underline{underlined}. Mistral 7B v0.1 also added for reference.}
\label{tab:detailed_scores}
\end{table*}

\end{document}